\pdfoutput=1

\documentclass[11pt]{article}

\usepackage[preprint]{acl}

\usepackage{times}
\usepackage{latexsym}

\usepackage[T1]{fontenc}

\usepackage[utf8]{inputenc}

\usepackage{microtype}

\usepackage{inconsolata}

\usepackage{booktabs}
\usepackage{amsmath}
\usepackage{amssymb}
\usepackage{graphicx}
\usepackage{pdfpages}
\usepackage{multirow}
\usepackage{subcaption}

\usepackage{xcolor}

\title{Symmetric Dot-Product Attention for Efficient Training\\ of BERT Language Models}

\author{
    Martin Courtois, Malte Ostendorff, Leonhard Hennig,  Georg Rehm \\
    Deutsches Forschungszentrum für Künstliche Intelligenz GmbH (DFKI), Berlin, Germany \\ Corresponding author: \texttt{martin.courtois@dfki.de}  \\
}

\begin{document}
\maketitle
\begin{abstract}

Initially introduced as a machine translation model, the Transformer architecture has now become the foundation for modern deep learning architecture, with applications in a wide range of fields, from computer vision to natural language processing. Nowadays, to tackle increasingly more complex tasks, Transformer-based models are stretched to enormous sizes, requiring increasingly larger training datasets, and unsustainable amount of compute resources. The ubiquitous nature of the Transformer and its core component, the attention mechanism, are thus prime targets for efficiency research.
In this work, we propose an alternative compatibility function for the self-attention mechanism introduced by the Transformer architecture. This compatibility function exploits an overlap in the learned representation of the traditional scaled dot-product attention, leading to a symmetric with pairwise coefficient dot-product attention. When applied to the pre-training of BERT-like models, this new symmetric attention mechanism reaches a score of $79.36$ on the GLUE benchmark against $78.74$ for the traditional implementation, leads to a reduction of $6\%$ in the number of trainable parameters, and reduces the number of training steps required before convergence by half.
\end{abstract}

\section{Introduction}

Since its introduction in 2017, the Transformer architecture powered by its scaled dot-product attention mechanism \citep{custom:vaswani2017attention} has become the core component of modern deep-learning architectures and has enabled researchers to achieve breakthroughs in both natural language processing (NLP) and computer vision tasks such as language modelling \citep{custom:brown2020language}, machine translation \citep{custom:raffel2019exploring}, speech processing \citep{custom:radford2022robust}, and image recognition \citep{custom:dosovitskiy2020image}. One of the many successes of the Transformer lies in its ability to operate and learn in an unsupervised setting from unstructured textual data, as well as its ability to handle complex and varied structures such as graphs, images, and sentences by increasing the model's number of layers. However, this trend has led to the emergence of machine learning models so enormous that the gap between the amount of compute resources available to many research groups and the amount needed to stay competitive is increasing year after year \citep{custom:togelius2023choose}, and by training larger and larger models, brought deep-learning's energy consumption to unsustainable amounts \citep{custom:thompson2021thecost}.

\begin{figure}[t!]
\centering
\includegraphics[width=.35\columnwidth]{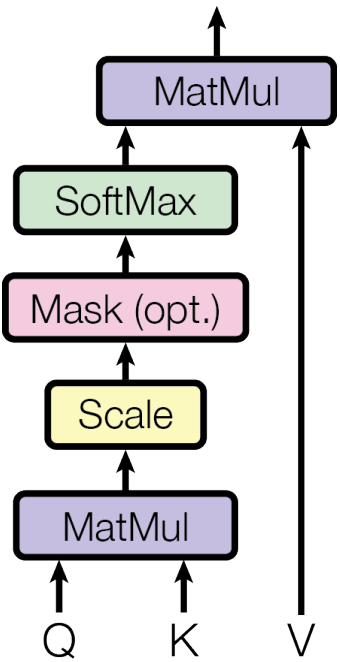}
\caption{Scaled Dot-Product Attention \citep{custom:vaswani2017attention}}
\label{fig:scaled-dot-product}
\end{figure}

Efficient Transformer implementations are a popular area of research with many recent contributions on encoding and dense representation of tokens \citep{custom:su2021rotary}, hardware-optimized implementation of attention \citep{custom:dao2022flashattention}, or implementations for long document processing \citep{custom:beltagy2020longformer}. While the attention mechanism itself has been studied extensively \citep{custom:niu2021review}, and several improvements to its computational complexity have been achieved \citep{custom:kitaev2020reformer,custom:zhou2021informer}, it is still primarily computed via the dot-product between a \emph{query} and a \emph{key} (see Figure~\ref{fig:scaled-dot-product}). \citet{custom:vaswani2017attention} highlight the difficulty of determining a proper compatibility function, and suggest that a more sophisticated compatibility function than dot product may be beneficial.

In this work, we propose alternative compatibility functions for the attention mechanism, i.\,e., the scaled dot-product attention mechanism. With this approach, we aim to improve the training efficiency of Transformer-based models and to reduce their resource consumption. We especially focus on the \emph{self-attention} mechanism of BERT \citep{custom:devlin2018bert}, a Transformer-based encoder model.

Our contributions can be summarized as follows:
\begin{itemize}
    \item We introduce an alternative formula to replace the scaled dot-product attention (Section~\ref{sec:symmetric-attention}) that takes advantage of the underlying symmetric structure of attention, in order to reduce the number of parameters and improve the computational efficiency of the model.
    \item We benchmark our approach by training several BERT models on three attention mechanism setups as well as two different model sizes (Section~\ref{sec:experiments}).
\end{itemize}
We demonstrate that our new attention formula reduces the number of parameters of the model by $6\%$, and achieves a reduction of the number of training steps required for model convergence by $50\%$ without sacrificing accuracy (Section~\ref{sec:results}).
Finally, we discuss the effects of our proposed compatibility function on training efficiency, and situate our approach within the context of research on efficient Transformer-based models (Section~\ref{sec:discussion}).

\section{Improving the Attention Mechanism}
\label{sec:symmetric-attention}

Modern Transformer-based models are neural networks that rely on the scaled dot-product attention mechanism introduced by \citet{custom:vaswani2017attention}. We propose two variations of this mechanism: a symmetric dot-product and a symmetric with pairwise factors dot-product, that lead to a reduction in the number of parameters of the self-attention layer.

\subsection{Scaled Dot-Product Attention}

The scaled dot-product attention given by the following equation (Equation~\ref{eqn:scaled-dot-product-attention}) is an operator on three input matrices, queries $Q$, keys $K$ and values $V$. We focus on the dot product $QK^T$ between queries $Q$ and keys $K$, which is responsible for measuring the compatibility between tokens. The compatibility $\mathbf{A}$ is an operator on two input tokens ($x, y \in \mathbb{R}^h$), that computes the dot-product of the projections of $x$ and $y$ respectively through the operators $\mathbf{Q}$ and $\mathbf{K}$:

\begin{equation}
\label{eqn:scaled-dot-product-attention}
Attn(Q, K, V)=\operatorname{softmax}\left(\frac{\mathbf{QK^{T}}}{\sqrt{d}}\right) \mathbf{V}
\end{equation}

Given two linear operators $\mathbf{Q}: \mathbb{R}^h \to \mathbb{R}^{d}$ and $\mathbf{K}: \mathbb{R}^h \to \mathbb{R}^{d}$, we define a compatibility operator $\mathbf{A}: \mathbb{R}^h \times \mathbb{R}^h \to \mathbb{R}$, such that:
\begin{equation}
    \mathbf{A}(x, y)=\mathbf{Q}(x) \cdot \mathbf{K}(y)^T
\label{eqn:affinity}
\end{equation}

We challenge the necessity of using two different operators to compute the affinity of the self-attention encoder layer of the Transformer block. Since both $\mathbf{Q}$ and $\mathbf{K}$ are operators on the same token space, it is reasonable to assume that the representations they learn share some features. In that case, since the original expression (Equation~\ref{eqn:scaled-dot-product-attention}) does not enforce any feature sharing, it may possess redundant parameters that will need to be learned twice.

We attempt to make this feature sharing property explicit in the compatibility operator expression, in order to remove redundant parameters, reduce overall model size, and improve convergence rate.

\subsection{Symmetric Dot-Product attention}

A simple way to make the feature sharing property explicit, is to enforce the following relation $\mathbf{Q} = \mathbf{K}$ between the two operators. This ensures that $\mathbf{Q}$ and $\mathbf{K}$ share features and results in the symmetric compatibility operator:

\begin{equation}
\mathbf{A}_{sym}(x, y)=\mathbf{Q}(x) \cdot \mathbf{Q}(y)^T
\label{eqn:symmetric-affinity}
\end{equation}

\subsection{Pairwise Dot-Product Attention}

One aspect that needs to be considered is the amount of features shared between the two operators. Complete overlap in terms of features may be detrimental to the overall performance of the attention mechanism, e.\,g., it could prevent the model to learn asymmetric relationships. Thus, we suggest the following compatibility operator where the amount of feature sharing is learned during training. To achieve this, we start with an operator $\mathbf{L}$ that will be shared, and we define operators $\mathbf{Q}$ and $\mathbf{K}$ as a composition of $\mathbf{L}$ with a base change, resulting in the following compatibility operator (Equation~\ref{eqn:weighted-dot-product}):

Given a linear operator $\mathbf{L}: \mathbb{R}^h \to \mathbb{R}^{d}$ and two square matrices $W_q, W_k \in \mathbb{R}^{d \times d}$, we define two linear operators $\mathbf{Q}: \mathbb{R}^h \to \mathbb{R}^{d}$ and $\mathbf{K}: \mathbb{R}^h \to \mathbb{R}^{d}$, such that:
\begin{equation}
    \begin{split}
    \mathbf{Q}(x)&=\mathbf{L}(x) \cdot W_q \\ 
    \mathbf{K}(x)&=\mathbf{L}(x) \cdot W_k \\
     \end{split}
\end{equation}

Let $S \in \mathbb{R}^{d \times d}$ be the product $S = W_q \cdot W_k^T$, we define a compatibility operator $\mathbf{A}: \mathbb{R}^h \times \mathbb{R}^h \to \mathbb{R}$, such that:

\begin{equation}
    \begin{split}
    \mathbf{A}(x, y)&=\mathbf{Q}(x) \cdot \mathbf{K}(y)^T \\
    \mathbf{A}(x, y)&=\mathbf{L}(x) \cdot W_q \cdot W_k^T \cdot {\mathbf{L}(y)}^T \\
    \mathbf{A}(x, y)&=\mathbf{L}(x) \cdot S \cdot {\mathbf{L}(y)}^T \\
    \end{split}
\label{eqn:weighted-dot-product}
\end{equation}
This operator can be interpreted as a weighted dot-product whose weights are stored in $S$, a matrix of pairwise factors. To make the expression consistent with the previously established expressions (Equation~\ref{eqn:affinity} and Equation~\ref{eqn:symmetric-affinity}), we relabel the $\mathbf{L}$ operator with the letter $\mathbf{Q}$, resulting in the following pairwise compatibility operator (Equation~\ref{eqn:pairwise-affinity}):

\begin{equation}
\mathbf{A}_{pair}(x, y)=\mathbf{Q}(x) \cdot S \cdot {\mathbf{Q}(y)}^T
\label{eqn:pairwise-affinity}
\end{equation}

\subsection{Parameter Count}

\begin{table}[t!]
\centering
\begin{tabular}{lll}
\toprule
\textbf{Function}  & \textbf{Expression} & \textbf{Parameters} \\ \midrule
original  & $\mathbf{Q}(x)\mathbf{K}(y)^T$          & $\mathcal{O}(3h^2)$          \\ 
symmetric & $\mathbf{Q}(x) \mathbf{Q}(y)^T$          & $\mathcal{O}(2h^2)$          \\ 
pairwise  & $\mathbf{Q}(x) S \mathbf{Q}(y)^T$          & $\mathcal{O}(2h^2 + h^2/n)$          \\ \bottomrule
\end{tabular}
\caption{Parameter count of the attention layer per compatibility function.}
\label{tab:params}
\end{table}

For a Transformer block of $n$ heads, with input size $h$ and attention size $d$, we give the parameter count formula for a complete block (with parameters from $Q$, $K$ and $V$). We note that most Transformer implementations impose $d = h / n$.

As shown in Table~\ref{tab:params}, the symmetric compatibility operator uses two thirds of the original number of parameters. For the pairwise compatibility operator, the parameter count also depends on the number of attention heads, it converges towards $2/3$ of the original number of parameters as the number of attention heads increases.

In this section, we introduced two alternative compatibility functions for the attention mechanism, a symmetric dot-product operator and a symmetric with pairwise factors dot-product operator. In the following sections we will refer to them respectively as the \emph{symmetric operator} and the \emph{pairwise operator}, we will refer to the traditional scaled dot-product operator as the \emph{original operator}.

\section{Experiments}
\label{sec:experiments}

\begin{table}[t!]
\setlength{\tabcolsep}{4pt} 
\centering

\begin{tabular}{lll}
\toprule
\textbf{Config}                       & \textbf{Operator}   & \textbf{Parameters}            \\ \midrule
\multirow{3}{*}{$BERT_{small}$} & original  & 28,795,194            \\ 
                            & symmetric & 27,744,570 (3.65\%)   \\ 
                            & pairwise  & 27,875,642 (3.19\%)   \\ 
                            \midrule
\multirow{3}{*}{$BERT_{base}$}  & original  & 109,514,298           \\ 
                            & symmetric & 102,427,194 (6.47\%)  \\ 
                            & pairwise  & 103,017,018 (5.93\%)  \\ 
                            \bottomrule
\end{tabular}
\caption{Parameter count per model configuration and compatibility function (relative amount of parameters saved compared to the original). $Bert_{small}$: nlayers: 4, nheads: 8, hidden size: 512, intermediate size: 2048. $Bert_{base}$: nlayers: 12, nheads: 12, hidden size: 768, intermediate size: 3072.}
\label{tab:model_sizes}
\end{table}

To evaluate the symmetric and pairwise operators against the original operator, we train and evaluate several Transformer-based encoder models, each using a different compatibility operator as part of the self-attention mechanism. The models are trained under the same conditions. First, we pre-train the models, because we want to measure the evaluation loss during training to see if our modifications have an impact on the training efficiency and the accuracy of the model. Then, we evaluate each model on the GLUE benchmark \citep{custom:wang2018glue} to evaluate the model's accuracy on relevant downstream tasks, such as, sentence acceptability \citep{custom:warstadt2018cola}, sentiment analysis \citep{custom:socher2013sst2}, sentence similarity \citep{custom:cer2017stsb}, and natural language inference \citep{custom:williams2018mnli,custom:rajpurkar2016qnli}. Finally, we select model checkpoints during training and evaluate those checkpoints on GLUE to measure the models' accuracy on downstream tasks during training.

\subsection{Pre-Training Dataset}
To pre-train our models, we select a subset of 30 million English documents from the OSCAR corpus \citep{custom:abadji-etal-2022-towards,custom:2022arXiv221210440J} by applying content quality filters (see Appendix~\ref{sec:appendix}). Using OSCAR data instead of the BookCorpus \citep{custom:zhu2015bookcorpus} and Wikipedia dumps is recommended for training BERT models \citep{custom:geiping2023cramming} and ensures that the amount of documents is large enough for single epoch training.

This training dataset is tokenized using the pre-trained \emph{bert-base-uncased} tokenizer \cite{custom:devlin2018bert} and sentences are aggregated into groups of 512 tokens. After tokenization, the resulting dataset contains 137 million training samples, 70 billion tokens and 10,000 test samples.

\subsection{Model Architectures}

We prepare three variations of the BERT model \citep{custom:devlin2018bert} using the original, the symmetric and the pairwise operators. We also train on two model sizes, \emph{bert-small} and \emph{bert-base}. 

As shown in Table~\ref{tab:model_sizes}, the symmetric and pairwise operators lead to significant reduction in the number of parameters, $3.65\%$ and $3.19\%$ for the \emph{bert-small} model, $6.47\%$ and $5.93\%$ for the \emph{bert-base} model.

In the following sections, we refer to a \emph{bert-base} model as $BERT_{base}$ when it uses the original operator, $BERT_{base,sym}$ or $BERT_{base,pair}$ when it uses the symmetric or pairwise operator respectively.

\subsection{Pre-Training Setup}

We follow the pre-training setup described by \citet{custom:devlin2018bert}.
The models are trained on a pure masked language modeling task with masking probability of 0.15 and batch size of 256 samples per training steps. Models are trained on 200,000 steps with a linear learning rate of $10^{-4}$ and learning rate warm-up during the first 10,000 steps. For the optimizer, we use Adam \citep{custom:kingma2014adam} with weight decay, $\beta_1=0.9$, $\beta_2=0.999$, $\epsilon=10^{-12}$, resulting in models pre-trained on 26 billion tokens. We measure evaluation cross-entropy loss during training to assess the training efficiency of our models.

\subsection{Benchmark Fine-Tuning Setup}

After pre-training, the models are fine-tuned and benchmarked on the GLUE dataset \citep{custom:wang2018glue} to assess their natural language understanding (NLU) capabilities. Each model is fine-tuned on the provided downstream task training dataset for 5 epochs, with a batch size of 16 and a linear learning rate of $1 \cdot 10^{-5}$. This benchmarking step is repeated on 5 downstream trials with different seeds. We measure individual task's scores, benchmark average and standard deviation across all trials. For each model, we measure: the combined F1 and accuracy on the Microsoft Research Paraphrase Corpus \textbf{\emph{mrcp}} \citep{custom:dolan2005mrpc}, Matthews correlation on the Corpus of Linguistic Acceptability \textbf{\emph{cola}} \citep{custom:warstadt2018cola}, matched and mis-matched accuracy on the Multi-Genre Natural Language Inference Corpus \textbf{\emph{mnli}} \citep{custom:williams2018mnli}, accuracy on the Quora Question Pairs dataset \textbf{\emph{qqp}}\footnote{\url{https://data.quora.com/First-Quora-Dataset-Release-Question-Pairs}}, accuracy on the Recognizing Textual Entailment dataset \textbf{\emph{rte}} \citep{custom:dagan2006rte1,custom:bar2006rte2,custom:giampiccolo2007rte3,custom:bentivogli2009rte5}, the combined Pearson and Spearman correlation on the Semantic Textual Similarity Benchmark \textbf{\emph{stsb}} \citep{custom:cer2017stsb}, accuracy on the Stanford Question Answering Dataset \textbf{\emph{qnli}} \citep{custom:rajpurkar2016qnli}, and accuracy on the Stanford Sentiment Treebank \textbf{\emph{sst2}} \citep{custom:socher2013sst2}. The Winograd schema challenge \textbf{\emph{wnli}} task has been excluded from the evaluation following the recommendation of \citet{custom:devlin2018bert}.

Compared to the original BERT setup or more recent compute optimized fine-tuning setups \citep{custom:geiping2023cramming}, we choose to fine-tune for a longer time (5 epochs instead of 3) and with a lower learning rate ($1 \cdot 10^{-5}$ instead of $4 \cdot 10^{-5}$), to have a more stable fine-tuning experience and reduce the risk of lucky seeding. With this choice, we aim to have a fairer evaluation of the models.

\subsection{Checkpoint Benchmarking}

We want to evaluate how downstream accuracy evolves during pre-training. We extract checkpoints during training and evaluate them on the GLUE benchmark. Each checkpoint is fine-tuned and evaluated on GLUE using the previously established fine-tuning setup.

\section{Results}
\label{sec:results}

In this section, we present the results of our experiments, the pre-training of our three variants (Figure~\ref{fig:bert-training}), the scores they reach on the GLUE benchmark (Table~\ref{tab:glue_results}) once fully trained and the evolution of the GLUE score during training (Figure~\ref{fig:glue_checkpoints}).

\subsection{Pre-Training Experiment}

\begin{figure*}[t!]
  \begin{subfigure}[b]{0.48\textwidth}
    \includegraphics[width=\columnwidth]{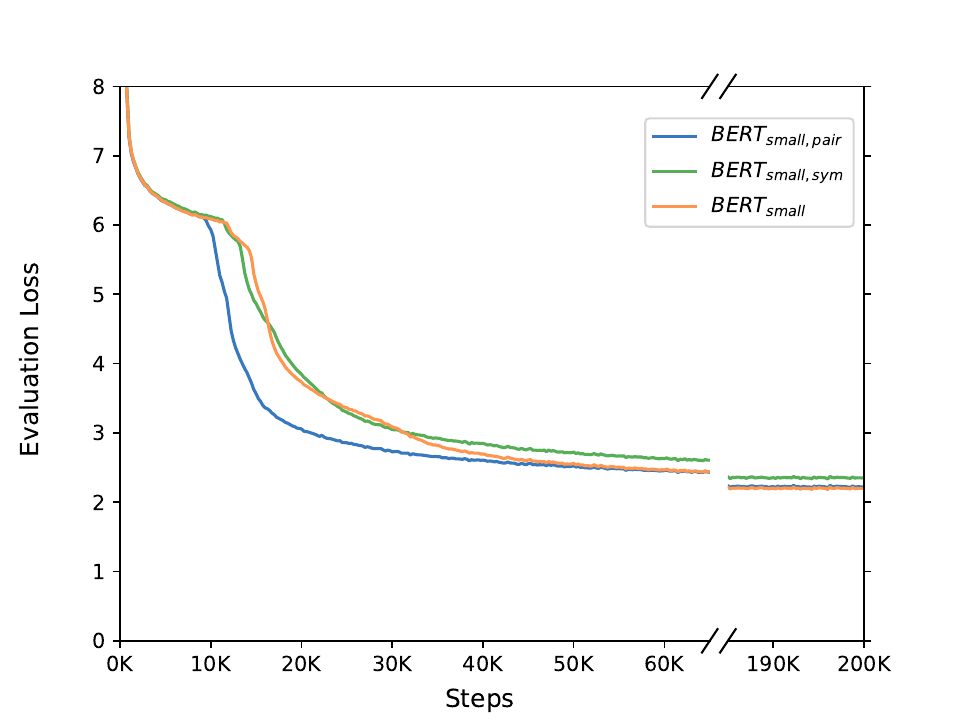}
    \caption{$BERT_{small}$}
    \label{fig:bert-small-training}
  \end{subfigure}
  \hfill
  \begin{subfigure}[b]{0.48\textwidth}
    \includegraphics[width=\columnwidth]{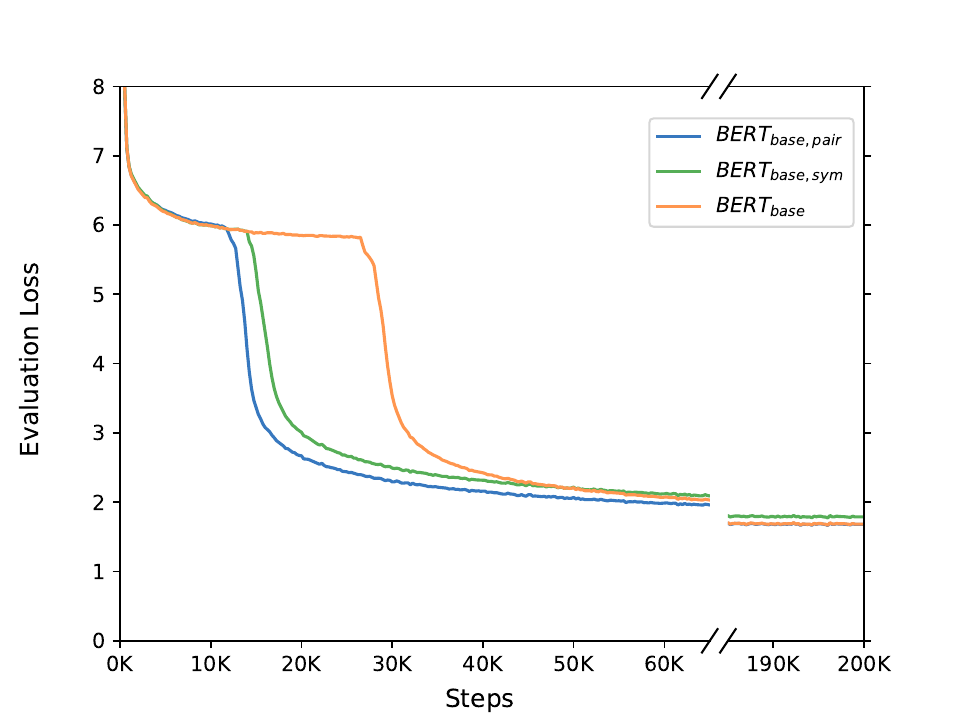}
    \caption{$BERT_{base}$}
    \label{fig:bert-base-training}
  \end{subfigure}
  \caption{BERT pre-training evaluation loss. Models are trained for 200,000 steps, the evaluation loss is the cross-entropy loss. We observe that the models using the symmetric and pairwise operators converge faster than the original model.}
  \label{fig:bert-training}
\end{figure*}

Figure~\ref{fig:bert-base-training} shows that the symmetric and pairwise variant converge much faster than the original variant for the $BERT_{base}$ model. The evaluation loss of the original variant remains on the initial plateau until step 25,000, when it sharply decreases. The symmetric variant remains on the initial plateau until step 13,500 and the pairwise variant until step 12,000. We also note that the original and pairwise variants will eventually reach the same evaluation loss plateau, while the symmetric variant remains above the two other variants with an additional absolute error of $0.1$.

Comparing Figures~\ref{fig:bert-small-training} and~\ref{fig:bert-base-training}, we observe the impact of model size on training efficiency. When the model size increases, the original variant's initial plateau is expanded from step 12,000 to step 25,000, while the symmetric and pairwise variant were almost unaffected.

\subsection{GLUE Benchmark Fine-Tuning}

Table~\ref{tab:glue_results} shows that the pairwise variant performs better than the original variant with an increase of $0.6$ points on the average GLUE score for both model sizes. The symmetric variant, however, is outperformed by the original variant in both cases, with a drop of $4$ points on the average GLUE score. We also observe that both proposed variants have a lower standard deviation on the \emph{bert-base} model.

\subsection{GLUE Benchmarking Along Training Steps}

\begin{figure}[t!]
\includegraphics[width=\columnwidth]{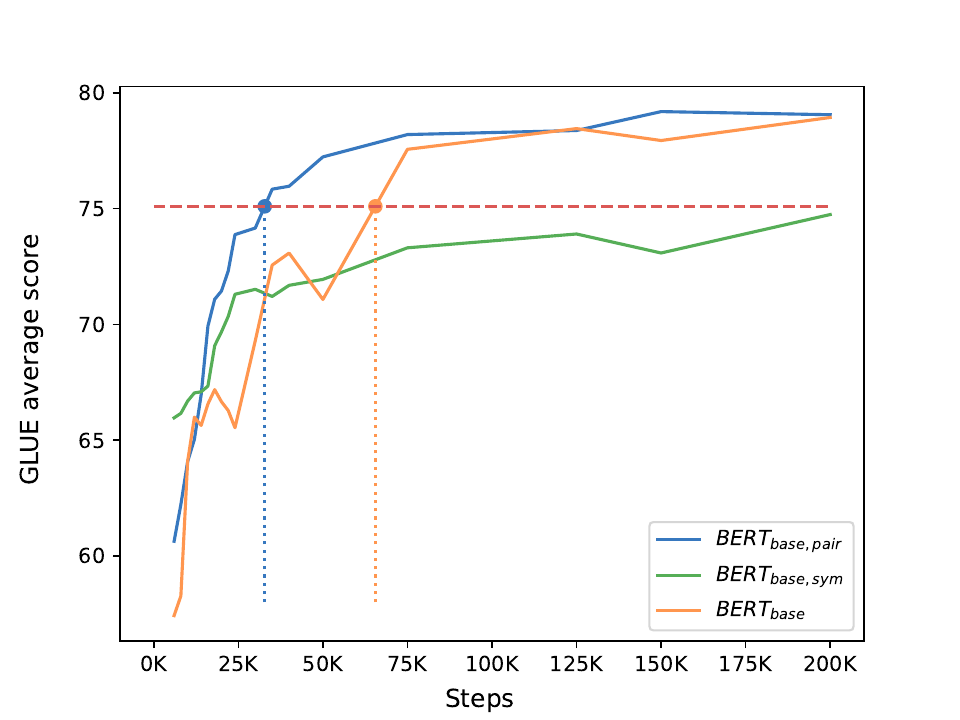}
\caption{Average GLUE score over training steps. Checkpoints are sampled during training and evaluated on the GLUE benchmark. The red dashed line correspond to $95\%$ of the final GLUE average score.}
\label{fig:glue_checkpoints}
\end{figure}

Figure~\ref{fig:glue_checkpoints} shows that the improved training efficiency observed during pre-training translates to a faster convergence rate on the GLUE benchmark as well. The pairwise and original variants both reach a final average GLUE score of approximately $79$. The pairwise variant achieves $95\%$ (a score of $75$) of its final value after 30,000 steps, the original variant reaches the same score after 65,000 steps.

We also observe a smoother evolution of the accuracy for the pairwise variant compared to the original variant. The experiment also highlights the performance drop of the symmetric variant when compared to the original variant.

\section{Discussion}
\label{sec:discussion}

\begin{table*}[!htb]
\resizebox{\textwidth}{!}{%
\begin{tabular}{lrrrrrrrrr}
\toprule
\textbf{Model}                    & \multicolumn{1}{l}{\textbf{GLUE Score}}        & \multicolumn{1}{l}{\textbf{mrpc}} & \multicolumn{1}{l}{\textbf{cola}} & \multicolumn{1}{l}{\textbf{mnli(m/mm)}} & \multicolumn{1}{l}{\textbf{qqp}} & \multicolumn{1}{l}{\textbf{rte}} & \multicolumn{1}{l}{\textbf{stsb}} & \multicolumn{1}{l}{\textbf{qnli}} & \multicolumn{1}{l}{\textbf{sst2}} \\ \midrule
$BERT_{small}$               & 72.72 (0.07) & 81.04                     & 21.39                     & \textbf{77.16/77.76}            & 86.08                    & 54.87                    & 82.20                     & \textbf{85.45}            & 88.49                     \\ 
$BERT_{small,sym}$             & 69.61 (0.32) & 76.81                     & 10.29                     & 75.25/75.72                     & 85.13                    & 55.74                    & 77.83                     & 82.79                     & 86.90                     \\ 
\textbf{$BERT_{small,pair}$} & \textbf{73.38 (0.37)}             & \textbf{82.34}            & \textbf{24.21}            & 76.37/76.89                     & \textbf{86.67}           & \textbf{56.25}           & \textbf{84.13}            & 84.97                     & \textbf{88.60}            \\ \midrule
$BERT_{base}$                & 78.74 (0.63)                      & 85.30                     & 44.35                     & \textbf{81.66/82.07}            & 88.86                    & 59.42                    & \textbf{87.30}                     & 88.76                     & \textbf{90.92}            \\ 
$BERT_{base,sym}$            & 74.82 (0.36)                      & 78.36                     & 35.22                     & 78.66/79.05                     & 87.70                    & 53.43                    & 84.47                     & 86.90                     & 89.56                     \\ 
\textbf{$BERT_{base,pair}$}  & \textbf{79.36 (0.37)}             & \textbf{87.83}            & \textbf{46.91}            & 81.60/82.02                     & \textbf{88.89}           & \textbf{60.58}           & 86.88           & \textbf{88.78}            & 90.78                     \\ \bottomrule
    \end{tabular}
    } 
    \caption{Average GLUE scores over the GLUE benchmark per model with individual task breakdown. $BERT_{base,pair}$ achieves the best \textbf{GLUE Score} of 79.36 with a standard deviation of 0.37, in comparison to $BERT_{base}$ which achieve a \textbf{GLUE Score} of 78.74 with a standard deviation of 0.63.}
    \label{tab:glue_results}
\end{table*}

\subsection{Pre-training Efficiency}

During the pre-training experiment, we observed that both variants $BERT_{base,sym}$ and $BERT_{base,pair}$ outperformed the original variant $BERT_{base}$ in terms of convergence rate (they initiated the learning and reached their respective plateau faster), for a \emph{bert-base} model the convergence rate seems to be two times faster. However, $BERT_{base}$ and $BERT_{base,pair}$ ultimately met around the same evaluation loss, while $BERT_{base,sym}$ performed a little worse.

One obvious explanation for the improved convergence rate can be found in the reuse of the $\mathbf{Q}$ operator, this can impact convergence rate in three way:
\begin{itemize}
    \item The accumulation of two loss gradients per forward/backward pass instead of a single one, resulting in an effect similar (but not exactly equivalent) to doubling the learning rate for the parameters of the $\mathbf{Q}$ operator.
    \item The reduction in the number of parameters.
    \item Sharing representation for both $\mathbf{Q}$ and $\mathbf{K}$ operators. If they do learn a subset of the same features, then enforcing a shared representation for both of them will reduce the amount of learning required.
\end{itemize}
These effects explain why both $BERT_{base,sym}$ and $BERT_{base,pair}$ converge much faster than $BERT_{base}$.

While converging faster than $BERT_{base}$, $BERT_{base,sym}$ did not reach the same evaluation loss. It is fair to assume that this is a modelling issue and not a size issue since $BERT_{base,pair}$ outperformed $BERT_{base,sym}$ with a similar number of parameters. Thus, we can conclude that symmetry is not a desired property of the compatibility function of the attention mechanism.

\subsection{GLUE Benchmark}

The evaluation of the three variants on the GLUE benchmark shows that $BERT_{base,pair}$ is more accurate than $BERT_{base}$, reaching an average score of $79.39$ against $78.74$ respectively. The evaluation also shows that the standard deviation of the average score across five trials is lower for both $BERT_{base,pair}$ and $BERT_{base,sym}$, with a standard deviation of $0.37$ and $0.36$ against $0.63$ for $BERT_{base}$.

This confirms that the training efficiency improvement observed on the pre-training task translates to the fine-tuning task and leads to improvement on the downstream task's accuracy. With the added benefit of making the fine-tuning task more stable, as shown by the lower standard deviation.

We also note that the fairly small $0.1$ difference in evaluation loss during training for $BERT_{base,sym}$ has translated to a $4$ points accuracy drop on the evaluation benchmark, echoing our remark on the need to model asymmetric relationships.

With these results, we experimentally prove that our pairwise operator improves the training efficiency of Transformer-based models, leading to a faster convergence rate and overall lower training loss. These improvements also translate to downstream task benchmarks. Models using the pairwise compatibility operator are indeed more accurate than the ones using the original compatibility operator.

\subsection{GLUE Evaluation During Pre-Training}

Running the benchmark evaluation on our three models at several steps of the pre-training experiment shows that the training efficiency we observed translates well into downstream accuracy. Our $BERT_{base,sym}$ and $BERT_{base,pair}$ converge faster towards their respective final values, similarly to the training loss observed on the pre-training task. $BERT_{base}$ reaches $95\%$ of its final value after 65,000 steps and $BERT_{base,pair}$ after 30,000 steps. While $BERT_{base}$ eventually catches up and improves on $BERT_{base,sym}$, $BERT_{base,pair}$ is consistently the better model.

This final experiment highlights the improved training efficiency induced by the pairwise compatibility operator. The faster convergence rate observed during pre-training is also observed on the downstream task evaluation, confirming the convergence rate improvement by a factor of two for the $BERT_{base,pair}$ model.

\section{Related Work}
\label{sec:related work}

While the Transformer architecture \citep{custom:vaswani2017attention} popularized the use of the attention mechanism, and contributed to its adoption in the field of NLP, the attention mechanism was first introduced to NLP with recurrent neural networks applied to machine translation \citep{custom:bahdanau2016neural}. In this setting, the compatibility operator is a simple multi-layer perceptron with non-linear activation operating on the concatenation of inputs encoded by the recurrent neural network. This definition of the attention mechanism was then extended to other compatibility operator: \citet{custom:luong2015effective} mention the use of the dot-product between the recurrent neural network's hidden state, propose to explicitly integrate token positions into the compatibility operator, and even suggest the use of a general dot-product operator $score(h_t, \overline{h_s})=h_t^T W \overline{h_s}$. Those initial influences have also been documented by \citet{custom:galassi2020attention} and \citet{custom:niu2021review}, where the general dot-product appears as a weighted dot-product between query and keys $f(q, K)=q^T W K$. Thus the pairwise compatibility operator we introduce is an evolution of the general dot-product, where we constrain it to a single and shared linear operator $\mathbf{Q}$ before applying the bilinear form of matrix $S$, resulting in the following operator $\mathbf{A}(x, y) = \mathbf{Q}(x) S \mathbf{Q}(y)^T$.

To the best of our knowledge, our work is the first application of the general dot-product with enforced symmetry to the self-attention mechanism of the Transformer architecture. While we focused on the compatibility operator, recent improvements have been made on other parts of the attention mechanism. Namely, \citet{custom:he2023simplifying} proposed to simplify the entire Transformer block by carefully removing components and achieved an impressive $15\%$ weight reduction, while still relying on the traditional scaled dot-product.

\section{Conclusions}

In this work, we revisited the traditional scaled dot-product used in the Transformer self-attention mechanism. We challenged the use of two distinct operators to compute the dot-product between queries and keys, in favor of single shared operator and a weighted dot-product with pairwise factors. By doing so, we enforced a symmetric structure to the compatibility operator of the attention mechanism, reducing the number of parameters used in the Transformer layer by a third. As a result, when applied to BERT models, our pairwise compatibility operator reduces the overall number of parameters of the model by $6\%$, reduces the number of pre-training steps required by half and improves accuracy on the GLUE benchmark, making Transformer-based encoders more efficient, faster to train and lowering their resource requirements. We believe our work can be applied to other Transformer architectures like decoder and encoder-decoder models, as well as to other NLP tasks like machine translation and language modeling. And, more generally, to the concept of attention as a whole, where it would bring improvement in other fields such as computer vision.

For future work, we plan to evaluate the pairwise dot-product attention mechanism on larger models reaching into the billion parameters, and to evaluate our attention mechanism on other benchmarks, like SuperGLUE \citep{custom:wang2019superglue} and SQuAD2.0 \citep{custom:rajpurkar2018squad2}. We plan on implementing the pairwise compatibility operator for the cross-attention mechanism, and evaluating it on decoder and encoder-decoder tasks like language modeling and machine translation. Finally, we want to evaluate our pairwise dot-product attention not only on natural language processing tasks, but also on tasks from other fields, computer vision, time series forecasting and reinforcement learning.

\section*{Limitations}
\label{sec:limitations}

Our work focuses only on the application and evaluation of alternative compatibility functions for the self-attention mechanism of Transformer-based encoder models, benchmarked on NLU tasks. While our work has shown positive results on this specific use case, we cannot draw any conclusion on its application to decoder models and pure language modeling tasks, or encoder-decoder model and machine translation tasks. Those use cases rely on the cross-attention mechanism for which the shared representation we exploit with our pairwise compatibility operator may not be appropriate.

While we suggest that the $\mathbf{Q}$ and $\mathbf{K}$ operators learn a shared representation, we did not perform any analysis of the original scaled-dot product attention or of our pairwise dot-product attention. The parameter redundancy of multi-head attention models has been covered in \citet{custom:bian2021attention}. However, to our knowledge the parameter redundancy between the query and the key operator of a single head has not been studied.

While our work showed positive improvements on the training efficiency of BERT-like models of fairly small sizes (100 million parameters), it is not enough to draw conclusions on its efficiency on very large models (e.g., 10 billion parameters).

We decided to benchmark our models on GLUE, as it is the most popular benchmark for NLU evaluation. However, this benchmark as been largely surpassed by modern machine learning models. For that reason, new benchmarks have been introduced, such as SuperGLUE \citep{custom:wang2019superglue} or SQuAD2.0 \citep{custom:rajpurkar2018squad2}.

\section*{Reproducibility Statement}

All software related to our experiments with the attention mechanism is available online\footnote{\url{https://github.com/mcrts/ACL2024-SymmetricAttentionBert}}. It uses the PyTorch \citep{custom:paszke2017automatic} and Hugging Face Transformer \citep{custom:wolf2020transformers} frameworks. The necessary steps to recreate the training dataset are documented\footnote{\url{https://github.com/mcrts/ACL2024-SymmetricAttentionBert/tree/main/data}}, the dataset used for training is available on Hugging Face\footnote{\url{https://huggingface.co/datasets/mcrts/OSCAR-2301_en_30M}}.

\section*{Acknowledgements}
\label{sec:acknowledgements}
The work presented in this paper has received funding from the German Federal Ministry for Economic Affairs and Climate Action (BMWK) through the project OpenGPT-X (project no.~68GX21007D), and has been supported by the German Federal Ministry of Education and Research as part of the project TRAILS.

\bibliography{acl_latex}

\newpage
\appendix
\section{OSCAR Filters}
\label{sec:appendix}

To ensure high quality of our training dataset, we filter OSCAR dumps with the following rules:
\begin{itemize}
    \item From the UT1 Blocklists project\footnote{\url{http://dsi.ut-capitole.fr/blacklists/index_en.php}}, we exclude the following categories:
        \begin{itemize}
            \item ``agressif''
            \item ``adult''
            \item ``cryptojacking''
            \item ``dangerous\_material''
            \item ``phishing''
            \item ``warez''
            \item ``ddos''
            \item ``hacking''
            \item ``malware''
            \item ``mixed\_adult''
            \item ``sect''
        \end{itemize}
    \item We exclude documents whose \emph{harmful perplexity score} is below 5.0 and above 100,000.
    \item Following recommendation from \citet{custom:abadji-etal-2022-towards}, we exclude documents which have been flagged with quality warnings.
\end{itemize}

\end{document}